\documentclass{article}

\PassOptionsToPackage{numbers, compress}{natbib}

\usepackage[final]{neurips_2019}

\usepackage[utf8]{inputenc} 
\usepackage[T1]{fontenc}    
\usepackage{hyperref}       
\usepackage{url}            
\usepackage{booktabs}       
\usepackage{amsfonts}       
\usepackage{nicefrac}       
\usepackage{microtype}      
\usepackage{graphicx}

\title{Few Labeled Atlases are Necessary \protect\\ for Deep-Learning-Based Segmentation}

\author{%
  Hyeon Woo Lee\\
  BME\\
  Cornell University\\
   \And
  Mert R. Sabuncu \\
  ECE and BME\\
  Cornell University\\
  \And
  Adrian V. Dalca \\
  CSAIL, MIT\\
  MGH, HMS\\
}
\begin{document}

\maketitle

\begin{abstract}
We tackle biomedical image segmentation in the scenario of only a few \textit{labeled} brain MR images. This is an important and challenging task in medical applications, where manual annotations are time-consuming. Current multi-atlas based segmentation methods use image registration to warp segments from labeled images onto a new scan. In a different paradigm, supervised learning-based segmentation strategies have gained popularity. These method consistently use relatively large sets of labeled training data, and their behavior in the regime of a few labeled biomedical images has not been thoroughly evaluated.
In this work, we provide two important results for segmentation in the scenario where few labeled images are available. First, we propose a straightforward implementation of efficient  \textit{semi-supervised} learning-based registration method, which we showcase in a multi-atlas segmentation framework. Second, through an extensive empirical study, we evaluate the performance of a supervised segmentation approach, where the training images are augmented via random deformations. Surprisingly, we find that in both paradigms, accurate segmentation is generally possible even in the context of few labeled images.

\end{abstract}

%
%
%
%

\section{Introduction}
Biomedical image anatomical segmentation is a fundamental problem in medical image analysis. Recent state-of-the-art methods have focused on supervised deep learning based methods, which typically use large labeled datasets. However, acquiring a large set of paired manual segmentation maps is challenging and time consuming, leading to many datasets with \textit{few} labeled examples in practice. In this work, we investigate this scenario for brain MRI in common segmentation strategies. We show how different state of the art learning methods can yield impressive results with different properties in this setting, and analyze how the number of labeled atlases affect this result. We also propose a straightforward improvement that builds on these methods.

Multi-atlas segmentation (MAS) has been widely studied, especially in the context of only a few labeled images, or \textit{atlases}~\cite{MAS2009,mert_survey,koch2018,Wang2013}. To segment a new scan, and each atlas is first registered to the desired scan, the atlas label map is then propagated using the resulting deformation maps. Finally, warped labels from multiple atlases are fused to yield a final segmentation~\cite{sabuncu2010generative,mert_survey,koch2018,Wang2013}. For the first step, registration methods have traditionally solved an optimization problem for each image pair, and therefore exhibited long runtimes~\cite{ashburner2007,avants2008,bajcsy1989,beg2005computing,dalca2016patch,klein2009,krebs2019learning}. In contrast, recent learning-based registration approaches learn a function, usually a neural network, to take in two images and rapidly compute the deformation field. While some methods require many example (ground truth) deformations or segmentation maps~\cite{hu_media,yang2017quicksilver}, others are unsupervised, requiring only a dataset of images~\cite{dalca2019unsupervised}. Importantly, learning-based registration methods have fast runtimes, usually requiring only seconds for a registration at test time, even on a CPU~\cite{Balakrishnan2019,dalca2019unsupervised,devos2019}. Learning-based registration methods have been further extended to also leverage large labeled datasets at training to yield models that better align segmentation labels~\cite{Balakrishnan2019,hu_media}. In this paper, we build on this prior work in learning-based registration and propose a semi-supervised registration strategy that improves multi-atlas segmentation in this scenario of few available atlases.

Supervised learning based segmentation, especially using convolutional neural networks (CNNs), has recently seen tremendous success in segmentation~\cite{zeynettin2017,kamnitsas,kayalibay2017cnn,pereira2016,unet}. By seeing many examples, these methods learn the parameters of a CNN that takes an image as input and outputs a segmentation prediction. While these methods provide state-of-the-art results, they have generally been demonstrated in the context of large labeled datasets~\cite{zeynettin2017,kamnitsas}. Data augmentation strategies, such as random rotation, scaling, and smooth 3D deformations, are often performed to encourage robustness to image variations~\cite{zeynettin2017,hussian2018,vnet,unet}. We build on these methods and find that with careful data augmentation, good segmentations can be achieved even in our scenario of very few labeled training images.

There are several contemporaneous papers that are closely related to this paper. Some of these focus on a small (e.g. one) number of manually segmented images and leverage sophisticated data augmentations techniques or priors to facilitate supervised segmentation methods~\cite{chaitanya2019semi,zhao2019}. Others require no labeled examples of the desired modality, but exploit segmentation maps from other datasets~\cite{dalca2018anatomical,joyce2018deep}. In this paper, we focus on a comparative analysis of how different numbers of labeled data affect the performance of MAS and supervised approaches. Our experiment is carefully designed to understand when the use of each method is plausible or desired. Based on the insights we gain from our experiments, we propose a new semi-supervised method.

%
%
%
%

\section{Method}

Our goal is to segment a dataset of images, and we focus on brain MRI in our experiments. Let 
$\{I_i, S_i\}_{i=1}^N$ 
represent a small dataset of labeled \textit{atlases}, each consisting of the grayscale image~$I$ and discrete segmentation map~$S$, such that each voxel of~$S$ corresponds to one of L anatomical labels.

\subsection{Background}
\subsubsection{Image-Based Registration.} 

Let $I, I^*$ be an atlas and testing image, respectively. We build on learning-based registration methods that learn a model $g_\theta(I,I^*) = \phi$, where~$\phi$ is a registration field and~$\theta$ are parameters of the function, usually a convolutional neural network (CNN). 
The goal is for the network to yield deformations~$\phi$ such that for each voxel $p\in\Omega$, $I^*(p)$ and $[I \circ\phi](p)$ correspond to the same anatomical location, where~$I \circ \phi$ represents~$I$ warped by~$\phi$.
To optimize the parameters~$\theta$, supervised registration methods employ ``ground truth'' deformations that are either simulated or obtained using an external registration tool. To avoid the requirement of ground truth, we follow recent \textit{unsupervised} methods~\cite{Balakrishnan2019}. Specifically, we optimize network parameters $\theta$ using the loss

\begin{equation}
\label{eqn:reg-loss}
\mathcal{L}(\theta;I_i,I^*_j) = \mathcal{L}_{img}(I^*_j,I_i \circ g_\theta(I_i,I^*_j)) + \lambda\mathcal{L}_{smooth}(g_\theta(I_i,I^*_j)),
\end{equation}

using stochastic gradient descent where $\lambda$ is a regularization parameter, $I_i$ is a labeled atlas and $I^*_j$ is an image from a dataset of unlabeled images, $\{I^*_j\}$. $\mathcal{L}_{img}$ penalizes the dissimilarity between the image $I^*_j$ and the warped atlas $I_i \circ \phi$, while $\mathcal{L}_{smooth}$ encourages a smooth deformation.
We use normalized cross correlation (NCC) for $\mathcal{L}_{img}$, which has been shown to be robust to intensity heterogeneity, yielding better registration results than a simple loss such as mean square error~\cite{avants2008,Balakrishnan2019}.  

\vspace{-0.2cm}
\subsection{Semi-Supervised Registration}
To leverage the few existing atlas segmentation maps in a learning-based registration framework, we build upon the setup above and design a semi-supervised registration method. 
Specifically, during training with stochastic gradient descent, instead of always providing the network with a random atlas and an unlabeled image, we occasionally provide two atlases as input. In these instances, the network can be encouraged to \textit{also} optimize the accuracy of the segmentation overlap resulting from warping one atlas' label map using the resulting deformation $\phi$, building on recent label-supervised methods~\cite{Balakrishnan2019,hu_media}. 

Specifically, we add an additional segmentation term to the loss function:

\begin{eqnarray}
\label{eqn:loss2}
\mathcal{L}(\theta,I_i,I_j) & = & \mathcal{L}_{img}(I_j,I_i \circ g_\theta(I_j,I_i)) + \lambda\mathcal{L}_{smooth}(g_\theta(I_j,I_i))
\nonumber \\ & &+ \gamma\mathcal{L}_{seg}(S_{I_j}, S_{I_i}  \circ g_\theta(I_j,I_i)),
\end{eqnarray}

where $\gamma$ is a regularization parameter for the supervised loss, and $\mathcal{L}_{seg}$ captures the agreement of segmentation maps. Specifically, we employ the  Dice score, which has also been used in recent label-supervised registration methods in the context of large labeled datasets \cite{Balakrishnan2019,hu_media}. The Dice overlap of two atlases is then:
\begin{equation}
\label{eqn:dice}
\mathcal{L}_{seg}(S_{I_j}, S_{I_i} \circ\phi) = -\frac{1}{L}\sum_{l=1}^{L}\frac{2|S_{I_i}^{(l)}\circ\phi\cap S_{I_j}^{(l)}|}{|S_{I_i}^{(l)}\circ\phi|+|S_{I_j}^{(l)}|},
\end{equation}
for anatomical structures $l\in{L}$. This strategy leverages unlabeled images using equation~\ref{eqn:reg-loss}, and the few labeled images using equation~\ref{eqn:dice}, thus exploiting the topological consistency of anatomy offered by registration-based methods.

\subsection{Spatial Data Augmentation} \label{sda}
To train the CNN, at each iteration we randomly deform an atlas and its corresponding segmentation map $\{I, S_I\}$ with a smooth random deformation field
$\phi_r$:$\{\hat{I} =  I \circ\phi_r, \hat{S}_I\ = S_I\circ\phi_r\}$ as in segmentation methods~\cite{unet,zhao2019}. The warped segmentation $\hat{S}_I$ is only used in the semi-supervised loss~\ref{eqn:loss2}. As we demonstrate in our experiments, providing a synthesized atlas and segmentation map improves robustness of the learning based registration model, especially through iterations at which we register atlas to atlas during training. For supervised learning based segmentation strategy,  $\{\hat{I}, \hat{S}_I\}$ is used to train the CNN network.

\subsection{Multi-Atlas Segmentation}
Given a trained network, we warp labeled atlases and $N_I$ augmented atlases to each test subject. Rather than using nearest neighborhood interpolation, we propagate the segmentation probabilities encoded as one-hot matrix. 
We combine warped label probabilities by summing $N_I+N$ warped segmentation maps and the maximum likelihood label is assigned to each voxel.
\begin{figure}[t]
\includegraphics[width=\textwidth]{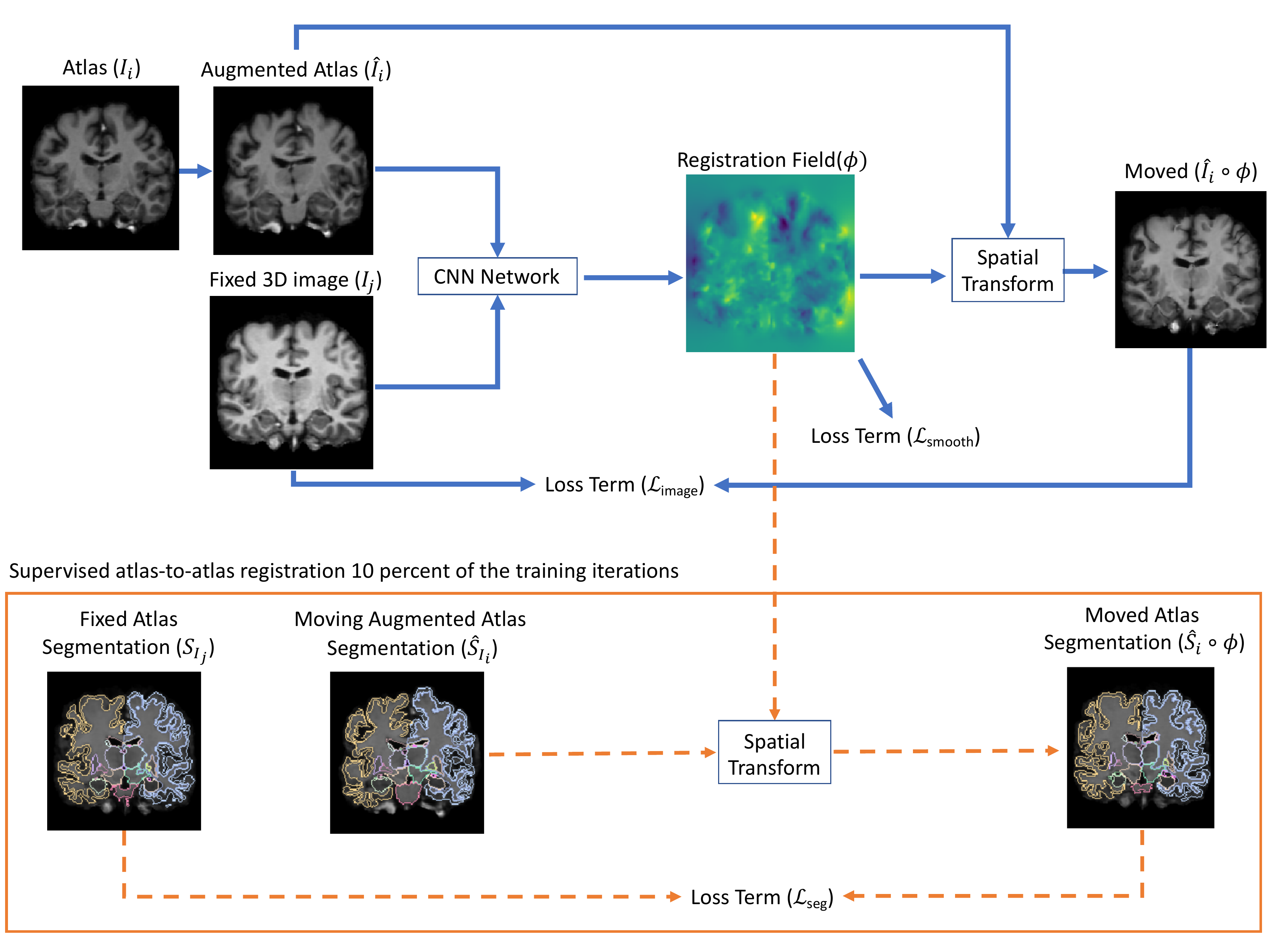}
\centering
\caption{Overview of end-to-end semi-supervised registration. For each training iteration, training image or atlas is augmented using a spatial deformation. We add optional semi-supervision (orange arrows) to the  unsupervised registration method. In our experiment setting, the supervised data is leveraged 10 percent of training iterations.}
 \label{fig:overall}
\end{figure}
\subsection{Implementation}
\begin{figure}[h]
\centering
\includegraphics[width=0.7\textwidth]{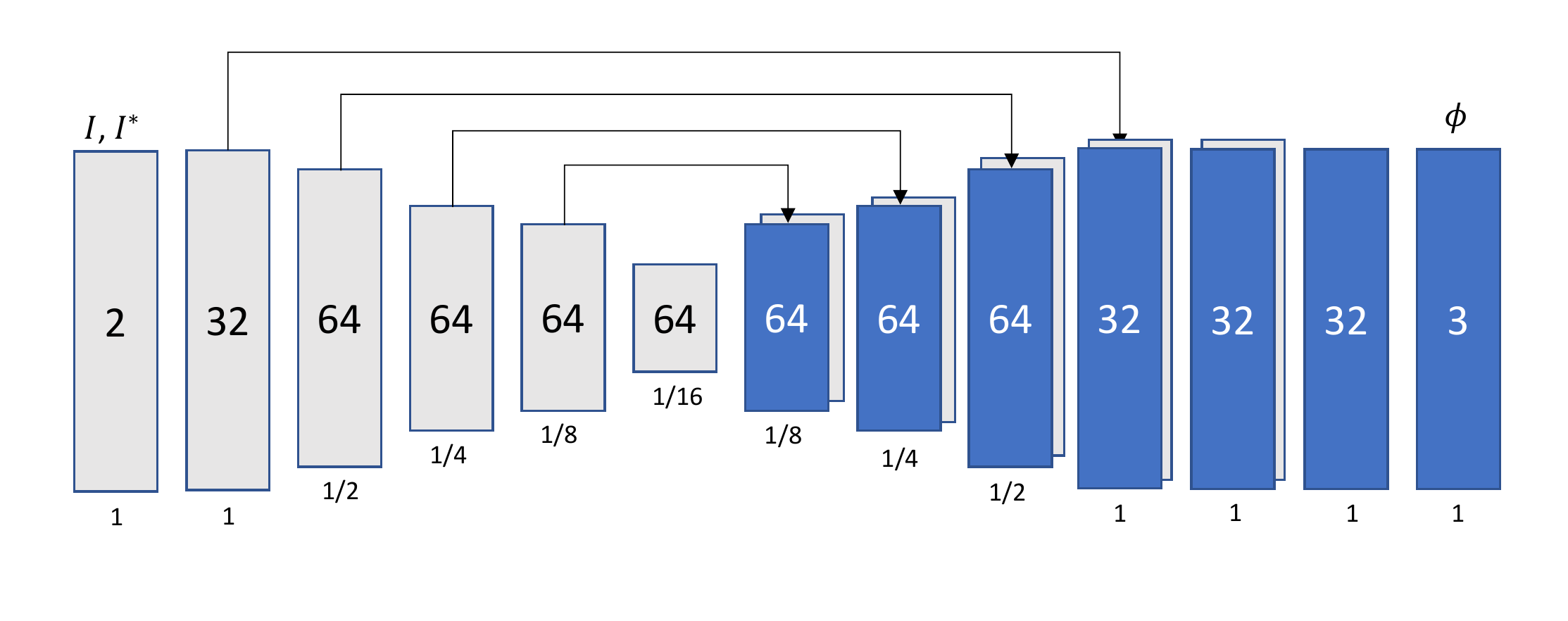}
\caption{\textbf{CNN architecture for image registration}. Each rectangle represents a 3D volume, the number inside the box indicates the number of filters, and the spatial resolution relative to the original size is included below each rectangle.
} \label{fig:cnn}
\end{figure}

We implement the registration function $g_\theta(\cdot,\cdot)$ as a CNN network with a UNet-style architecture following recent literature~\cite{Balakrishnan2019,unetSeg2017,kayalibay2017cnn}. Figure~\ref{fig:cnn} depicts the network used in registration. The network takes the 2-channel 3D image composed by concatenating the two inputs. Using a 3D UNet, we apply 3D convolution to the input with kernel size 3x3x3 and stride of 2, followed by Leaky ReLU activations. In the encoder, our convolution layers reduce a spatial dimension to half at each level. In the decoding stage, we use upsampling and convolution layers to increase the size of the image by 2 until the spatial dimension reaches to the original image size. Skip connections connect the respective layers in the encoder and decoder. We warp images using a spatial transformation function with linear interpolation. Figure~\ref{fig:overall} represents the overall pipeline of the proposed method. 

%

%
%

\section{Experiments}
We provide two main experiments with the goal of understanding the performance of models when few labeled atlases are available. First, we analyze the effect of our semi-supervised learning-based registration strategy on the performance of multi-atlas segmentation. Second, we analyze more broadly how MAS compares to supervised learning methods in the setting of few labeled examples. 

\subsection {Setup}
\subsubsection{Methods.}
We explore three variants of MAS with learning-based registration. MAS, MAS-DA, and MAS-SS refer to multi-atlas segmentation, multi-atlas segmentation with the proposed data augmentation (DA), and MAS with semi-supervised (SS) learning and DA, respectively. 

We also analyze supervised segmentation strategies.
Recent supervised learning-based segmentation methods use a discriminative CNN model~$f_\theta(I) = S$ that maps images I to their segmentation maps $S$ and is parametrized by $\theta$. We learn such a model using the labeled atlases, minimizing the categorical cross-entropy loss using stochastic gradient descent. Our focus is not to explore architecture variants but to compare this approach with MAS strategies. To preserve model capacity, we use the same UNet-style architecture as in the registration task.
We use softmax activation for the final layer to output the segmentation probabilities. For computational efficiency, we divide each image into 120 3D smaller patches of size 64x64x64. We train variants of supervised learning-based segmentation (which we call SegNet) with limited labeled data. SegNet-DA refers to a supervised method with data augmentation, implemented similar to section~\ref{sda}. Finally, as an upper bound, we train a fully supervised model, SegNet-Full, using the labels of all images (not just atlases) in the training set including segmentation maps not available to the other methods. We use this model simply to illustrate the optimal performance and enable a measure of the gap in performance compared to the rest of the tested models.

\subsubsection {Dataset.}
We use two datasets. First, we use preprocessed 7829 T1 weighted brain MRI scans from eight public data sets: ADNI \cite{adni}, OASIS \cite{oasis}, ABIDE \cite{abide}, ADHD200 \cite{adhd}, MCIC \cite{mcic}, PPMI \cite{ppmi}, HABS \cite{havard}, and Harvard GSP \cite{gsp}. All scans are preprocessed using FreeSurfer tools, including affine registration, brain extraction, and segmentation only used for evaluation~\cite{buckner}. 
We use 7329 random images from this dataset as unlabeled data for registration, and we emphasizes that labels from these dataset are not used during training the registration network. We similarly use a second dataset of 38 pairs of brain MRI scans and hand annotated segmentation maps from the Buckner40 dataset \cite{buckner}. We split this dataset into 18, 10, and 10 images for train, validation, and test sets, respectively. We train MAS models using atlases from Buckner40 training subset as input, and we use the eight public data sets as unlabeled data. For training SegNet models, we use atlases and corresponding segmentation maps from the Buckner40 train set. SegNet-Full refers to a SegNet-DA model that used all of the labeled data from Buckner40 and eight public dataset as training, and serves as upper bound model.

\subsubsection{Experimental Setup.} 
Our goal is to understand the behavior of MAS and supervised learning methods in the context of few labeled examples. We use $N=1..7$ labeled scans. Specifically, for each of $N=1...7$, $N$ atlases are randomly chosen from the Bucker40 training dataset. We repeat this process $n$ times to construct $n$ different random ``atlas sets''. For each MAS and supervised learning segmentation, $n$ models are trained and used to perform the segmentation on the test dataset. Evaluation of the performance for each paradigm is measured by averaging each evaluation metric (described below) over $n$ permutations. 

\subsubsection{Parameters.}
We set network architecture and parameters based on results in previous literature~\cite{Balakrishnan2019}. Specifically, we set regularization parameters $\lambda$ to 1.5 and $\gamma$ to 1.0. During training, we use the supervised atlas-to-atlas registration 10 percent of the time. On a small single scenario, we experimented with 50 percent and 10 percent and found that 10 percent produced optimal results.

\subsubsection{Evaluation Metric.}
We first evaluate our models with anatomical segmentation overlap using the Dice score \cite{Dice}. We focus on 29 anatomical structures that have significant volume in all images. The predicted segmentations are evaluated relative to manual anatomical segmentations from the Buckner40 dataset. Second, we evaluate surface distance (SD) of all structures. For each pre-defined anatomical region, we compute the distance between the predicted and manual segmentation surfaces in mm. SD is likely to highlight spurious segmentations which are further from correct edges. We average the metrics over structures and test subjects.

\begin{figure}
\includegraphics[width=\textwidth]{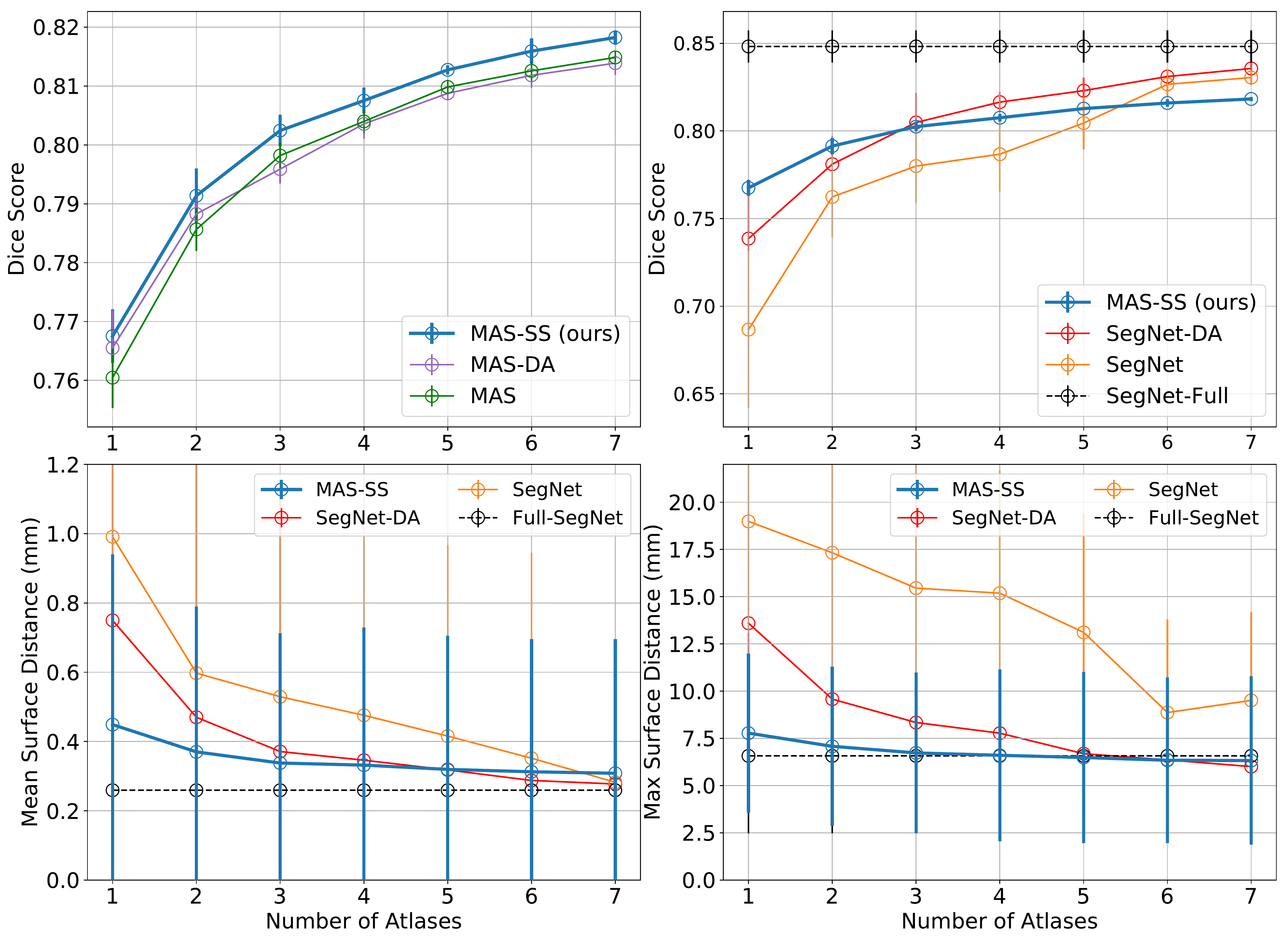}
\centering
\caption{Dice score and surface distance of test scans for various segmentation methods. Upper: mean Dice score (higher the better) for variants of MAS (left), and variants of SegNets (right). Lower: surface distance (lower the better) for MAS-SS and variants of SegNets: mean (left) and maximum (right).}
 \label{total_comp}
\end{figure}

\subsection{Results}
Figure~\ref{total_comp} presents the performance of each strategy on the test dataset. Surprisingly, we find that all methods, when used in their best variant, can achieve reasonable segmentation results that approach the upper bound demonstrated by the fully supervised SegNet model. For example, we find that with just three atlases, the best MAS and SegNet methods can be within four Dice points of this upper bound. Furthermore, our proposed method, MAS-SS, can acheive a maximum surface distance across all structures of less than 7 mm, and a mean surface distance less than 0.4 mm. 

We find that the registration data augmentation strategy improves the performance of MAS  in the few labeled atlases (less than three) setting in terms of Dice score. Importantly, our proposed model, MAS-SS, consistently improves on the performance of MAS and MAS-DA methods in terms of Dice score in all cases. All of the MAS models show consistently very small mean and max surface distance in Figure~\ref{total_comp}. 

Figure~\ref{total_comp} also illustrates segmentation performance of supervised learning-based strategies. Data augmentation significantly improves SegNet segmentation performance in terms of both Dice score and surface distance. Interestingly, MAS-SS yields higher Dice score for few atlases (less than three) and equivalent Dice score performance on three atlases. Importantly, the proposed method performs significantly better in terms of both mean and maximum surface distance given one to five or fewer atlases, highlighting the advantage of combining supervision with a registration-guided method that preserves anatomical topology. While SegNet-DA performs slightly better in terms of Dice score with more atlases, it still gives high surface distance. 

\begin{figure}
\includegraphics[width=\textwidth]{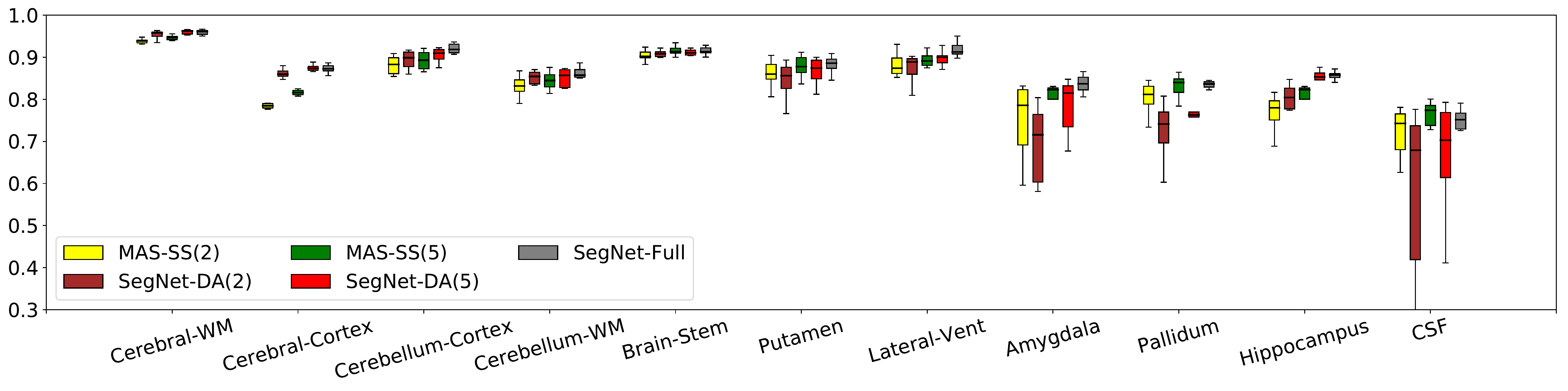}
\centering
\caption{Dice score of each segmentation method across various brain structures. We shorten white matter (WM), ventricle (vent), and cerebrospinal fluid (CSF). The number of atlases used for training is in parentheses next to each model in the legend.}
 \label{substructure}
\end{figure}
\begin{figure}
\includegraphics[width=\textwidth]{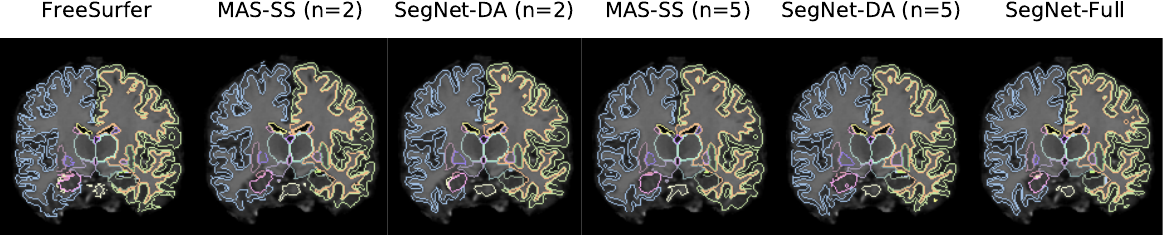}
\centering
\caption{Examples of MR slices with segmentation for several test subjects. We indicate the numbers of labeled atlases $n$ used for each method. 
} \label{slices}
\end{figure}

Figure~\ref{substructure} shows Dice scores for different brain structures. All methods achieve high Dice score for large volume structures, but MAS-SS consistently shows higher Dice score for small volume structures such as Amygdala, or Hippocampus. Figure~\ref{slices} shows example segmentation results from top performing strategies for mid-coronal slice. 

\section{Conclusion}
We  focus  on  segmentation  in  the  regime  of  few  labeled  data. Through a detailed comparison, we show the surprising result that even with few labeled images, two separate deep learning based approaches can achieve reasonable results, contrasting  conventional wisdom that deep learning approaches require many labeled scans. After investigating various state-of-art segmentation strategies with few labeled data, we also propose a semi-supervised, learning-based multi-atlas segmentation, which improves on existing methods. The proposed  method  achieves  both  high  Dice score improvement, but also low surface distance especially in context of few atlases. Furthermore, compared to supervised-learning based segmentation, the proposed method yields higher Dice score on anatomical structure of small volume, highlighting the advantage of a semi-supervised  framework  within  the  topologically-constrained  registration setting. These findings suggest two important contributions: first, the conclusion that deep learning segmentation strategies do not always require large amounts of labeled training data, and second, the semi-supervised learning method provides an improved approach to multi-atlas segmentation.

\bibliography{mybib}
\bibliographystyle{plain}

\end{document}